\documentclass[preprint,12pt]{elsarticle}

\usepackage{amssymb}
\usepackage{amsmath,amssymb,amsfonts}
\usepackage{booktabs,makecell,multirow}

\journal{Biomedical Signal Processing and Control}

\begin{document}
\begin{frontmatter}

\title{Score-based Generative Priors Guided Model-driven Network for MRI Reconstruction}

\cortext[cor1]{Corresponding author}
\affiliation[label1]{organization={Chongqing Key Laboratory of Image Cognition, Chongqing University of Posts and Telecommunications},
	city={Chongqing},
	country={China}}
\affiliation[label2]{organization={Key Laboratory of Cyberspace Big Data Intelligent Security, Chongqing University of Posts and Telecommunications, Ministry of Education},
	city={Chongqing},
	country={China}}

\author[label1]{Xiaoyu Qiao}\ead{d210201019@stu.cqupt.edu.cn}
\author[label1,label2]{Weisheng Li\corref{cor1}}\ead{liws@cqupt.edu.cn}
\author[label1]{Bin Xiao}\ead{xiaobin@cqupt.edu.cn}
\author[label1]{Yuping Huang}\ead{s180231053@stu.cqupt.edu.cn}
\author[label1]{Lijian Yang}\ead{eeejyang@gmail.com}

\begin{abstract}
Score matching with Langevin dynamics (SMLD) method has been successfully applied to accelerated MRI. However, the hyperparameters in the sampling process require subtle tuning, otherwise the results can be severely corrupted by hallucination artifacts, especially with out-of-distribution test data. To address the limitations, we proposed a novel workflow where naive SMLD samples serve as additional priors to guide model-driven network training. First, we adopted a pretrained score network to generate samples as preliminary guidance images (PGI), obviating the need for network retraining, parameter tuning and in-distribution test data. Although PGIs are corrupted by hallucination artifacts, we believe they can provide extra information through effective denoising steps to facilitate reconstruction. Therefore, we designed a denoising module (DM) in the second step to coarsely eliminate artifacts and noises from PGIs. The features are extracted from a score-based information extractor (SIE) and a cross-domain information extractor (CIE), which directly map to the noise patterns. Third, we designed a model-driven network guided by denoised PGIs (DGIs) to further recover fine details. DGIs are densely connected with intermediate reconstructions in each cascade to enrich the information and are periodically updated to provide more accurate guidance. Our experiments on different datasets reveal that despite the low average quality of PGIs, the proposed workflow can effectively extract valuable information to guide the network training, even with severely reduced training data and sampling steps. Our method outperforms other cutting-edge techniques by effectively mitigating hallucination artifacts, yielding robust and high-quality reconstruction results.
\end{abstract}


\end{frontmatter}

\section{Introduction}
\label{sec:introduction}
Magnetic Resonance Imaging (MRI) inherently requires long scanning time, leading to patient discomfort and degradation of image quality.
Undersampling the k-space data can proportionally reduce the scanning time but results in aliasing artifacts in the reconstructed images.
The conventional methods for reconstructing high-quality images from a small subset of k-space data are mainly based on parallel imaging (PI) \cite{larkman2007parallel, pruessmann2006encoding} and compressed sensing (CS) \cite{lustig2007sparse, lustig2008compressed, ye2019compressed}.
In recent years, deep-learning (DL)-based methods have been successfully introduced in the field of MRI reconstruction and have shown convincing improvements over conventional methods \cite{sandino2020compressed,liang2020deep}.
Early networks are designed in an end-to-end manner to fit the mapping between the input measurements and reconstruction targets \cite{zbontar2018fastmri, eo2018kiki}.
However, these methods still present problems such as lack of interpretability, and a relatively large dataset is required to fit the mapping.
Recently, some studies have focused on distribution-learning-based reconstruction processes using diffusion models \cite{kazerouni2023diffusion, korkmaz2023self, chung2022score, quan2021homotopic, wu2023wavelet, ozturkler2023smrd, cao2024high}.
Among them, denoising diffusion probabilistic models (DDPMs) \cite{ho2020denoising, wang2024srfs} and score matching \cite{liu2016kernelized,hyvarinen2005estimation} with Langevin dynamics (SMLD) methods, which correspond to variance preserving (VP) stochastic differential equations (SDEs) and variance exploding (VE) SDEs, respectively \cite{song2020score}, have shown great potential in solving inverse problems.
They have also been successfully extended to the MRI reconstruction field \cite{song2019generative, song2020improved, song2020score, jalal2021robust}.
However, the training (\textit{i.e.,} forward) and sampling (\textit{i.e.,} reverse) processes require careful tuning, otherwise the sampling results can be corrupted by hallucination artifacts, particularly with out-of-distribution test data.

\begin{figure}[!t]	
	\begin{minipage}[b]{1\linewidth}
		\centering        
		\centerline{\includegraphics[width=13cm]{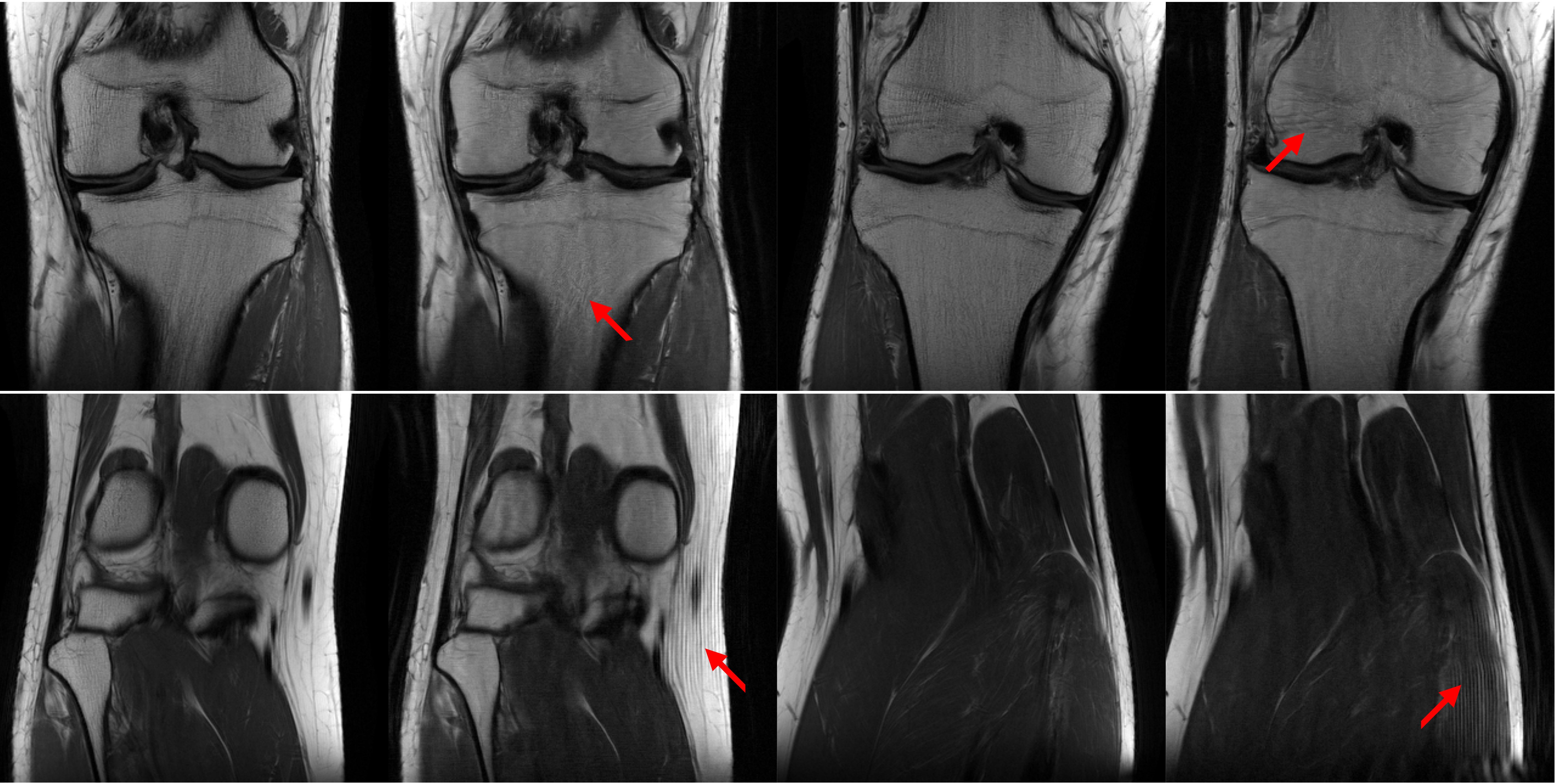}}
	\end{minipage}
	\caption{Distribution-shift reconstruction results of a pretrained score network on a knee MRI dataset. The score network was trained on a brain MRI dataset. Images in odd columns are ground truth and those in even columns are sampling results. \label{fig:1}}
\end{figure}

Model-driven networks are built in a different manner: conventional optimization algorithms are unrolled and realized in deep networks, and stacked cascades are used to mimic the iterative steps in the optimization procedure.
Compared with the other methods, unrolled networks offer better interpretability.
Enhanced reconstruction performance is observed with learning based parameters and regularizers in unrolled networks compared with traditional optimization procedure.
Although stacking more cascades can improve the reconstruction performance, large-scale backbone networks can hardly be applied to unrolled networks due to the limitations of GPU memory, and a trade-off must be established. Considering that distribution-learning-based and model-driven methods attempt to recover the same target images along different paths, the complementarity between these methods should be considered.
Current score-matching networks require retraining to achieve superior results across varying datasets.
We observed artifacts on the sampled images without network retraining and tuning of hyperparameters, as shown in Fig.~\ref{fig:1}. 
Despite these artifacts, we can observe that the artifact-affected images still provide information additional to the known undersampled measurements.
Therefore, if the artifacts can be appropriately eliminated, naive samples from SMLD methods can compensate for missing measurements and provide complementary information to guide model-driven network training.

To this end, instead of focusing on optimizing the forward and reverse process in SMLD methods, we aimed to design a model-driven network which can extract valuable information from naive sampling process and achieve high-quality results robustly.
We proposed a novel DL-based hybrid MRI reconstruction workflow, consisting of the following three main steps: 
(1) Sampling: Using known undersampled raw k-space data, we first obtained initial samples using SMLD methods.
We bypassed all the tuning steps in SMLD methods, such as tuning the hyperparameters or retraining the score networks on our dataset.
Instead, we adopted a pretrained score network (PSN) and used the naive samples as preliminary guidance images (PGI) for the follow-up steps.
A PSN can be trained using out-of-distribution data.
(2) Denoising: We believe that these preliminary reconstructions can provide additional information to guide the reconstruction process through appropriate denoising, leading to better performance.
Considering that the PGIs have relatively low quality and we observed similar artifacts in the images, we designed a denoising module (DM) to remove the artifacts and noises from the PGIs.
We designed a score-based information extractor (SIE) and a cross-domain information extractor (CIE) to build mappings from extracted features to the noise patterns.
Then, we used denoised PGIs (DGIs) as higher-quality guidelines for further reconstruction.
(3) Guided reconstruction: We designed a densely connected unrolled network to reconstruct the MR images from the undersampled measurements.
DGIs are simultaneously input with measurements and periodically updated in the network to guide network training.
The DM and unrolled cascades were trained end-to-end.
Finally, the proposed workflow essentially reconstruct MR images in a coarse-to-fine manner.
The denoising steps primarily restore low frequency patterns, while the guided reconstruction steps further recover fine details, achieving superior image quality.

The experimental results show that although the average quality of PGIs is even lower than that of the undersampled measurements, the artifacts were effectively eliminated by the DMs.
Meanwhile, the score network can not only help in the sampling process but also cooperate with the unrolled network and improve the denoising performance of the DMs.
Finally, the proposed reconstruction workflow outperformed state-of-the-art methods through the guidance of DGIs.

The main contributions of this study are summarized as follows:
\begin{itemize}
	\item We propose a novel DL-based MRI reconstruction workflow, and low-quality samples obtained using naive SMLD method can be leveraged as guidance during reconstruction.
	\item We propose a DM with SIE and CIE blocks which can coarsely remove artifacts and noises in the SMLD samples, and an unrolled network is proposed to further recover fine details.
	\item Experimental results demonstrate that without tuning the sampling steps or retraining the score network on target dataset, the low-quality samples can guide the proposed network to obtain robust and high-quality reconstructions, which outperforms the cutting-edge methods.
\end{itemize}

\section{Related Works}
\label{sec:rw}
For PI methods, multiple receiver coils are equipped to obtain raw k-space data simultaneously, and the missing data are recovered by exploiting redundancies among the data from different coils.
Sensitivity encoding (SENSE)-type \cite{pruessmann1999sense} methods work in the image domain, whereas some others directly interpolate missing data points in k-space, such as simultaneous acquisition of spatial harmonics (SMASH) \cite{sodickson1997simultaneous} and generalized autocalibration partial parallel acquisition (GRAPPA) \cite{griswold2002generalized}.
However, the reconstructions in PI approaches deteriorate at high acceleration factors.
In CS, sparsity-based priors are routinely incorporated to narrow the solution space in the objective functions, and iterative optimization algorithms are designed to approach artifact-free images \cite{block2007undersampled, feng2013highly, knoll2011second, fessler2020optimization, ramani2010parallel, huang2011efficient}.
However, conventional methods depend heavily on handcrafted procedures, and the iterative steps are time-consuming.

For deep networks trained end-to-end, the mapping between the undersampled measurements and reconstructed images are established.
For example, early studies adopted deep convolutional neural networks (CNNs) to reconstruct MR images from zero-filled inputs \cite{wang2016accelerating, ye2018deep}.
The authors in \cite{lee2018deep} trained the magnitude and phase networks separately in a deep residual network to remove aliasing artifacts.
In \cite{wang2020deepcomplexmri}, a complex-valued deep residual network with a higher performance and smaller size was proposed.
In addition to realizing image domain end-to-end learning, some studies have proposed direct learning mapping in k-space \cite{han2019k} or exploiting the complementarity between the image and k-space domains \cite{eo2018kiki, sheng2024cascade}. In \cite{li2024progressive}, the authors proposed an unsupervised framework which learn from image and frequency domain in a multi-stage manner.

Some other methods direct learned an image prior in generative models such as generative adversarial networks (GAN) \cite{yang2017dagan, mardani2018deep, quan2018compressed, zhou2021efficient, noor2024dlgan, sangeetha2024c2}.
For example, in \cite{bora2017compressed} the authors used pre-trained generative models to present data distributions.
More recently, diffusion models have emerged as new state-of-the-art generative models.
In \cite{song2019generative}, the authors used score matching to estimate the data distribution, and samples were then generated using annealed Langevin dynamics.
Subsequently, the authors proposed improved skills in \cite{song2020improved} for higher stability and better performance, and exact log-likelihood computation was supported in \cite{song2020score}.
\cite{jalal2021robust} first successfully applied CS-based generative models to overcome the MRI reconstruction problem.
In \cite{ozturkler2023smrd}, the authors further optimized the sampling stage by automatically tuning the hyperparameters using Stein's unbiased risk estimator (SURE).
The authors in \cite{cao2024high} proposed a high-frenquency-based diffusion process and accelerated the reverse process.
A wavelet-improved technique was proposed in \cite{wu2023wavelet} to stably train the score network.
In the reverse process, the authors further designed regularization constraints to enhance the robustness.
Some works extended the idea of cold diffusion \cite{bansal2024cold} to MRI reconstruction, and the process of adding Gaussian noise is replaced with undersampling operation \cite{huang2023cdiffmr}.
However, these networks generally required large training datasets \cite{wang2021deep} and suffer from the lack of interpretability.
Moreover, tuning the training and sampling process of diffusion models is non-trivial and crucial for achieving robust reconstruction outcomes. 

Unrolled networks, also known as model-driven networks, are constructed by unrolling iterative optimization steps into deep networks \cite{sun2016deep, zhang2018ista, hammernik2018learning, duan2019vs,sriram2020end, zhao2024j, wang2024progressive}. For example, algorithms such as the alternating direction method of multipliers (ADMM) \cite{sun2016deep}, iterative shrinkage-thresholding algorithm (ISTA) \cite{zhang2018ista}, variable splitting (VS) \cite{duan2019vs}, and approximate message passing (AMP) have been successfully applied to solve the objective functions in MRI reconstruction. 
In \cite{sriram2020end}, the authors extended a previous variational network \cite{hammernik2018learning} to learn end-to-end, and sensitivity maps (SM) were estimated in the network.
To further improve the accuracy of the SM, different studies \cite{jun2021joint, arvinte2021deep} have explored ways to simultaneously update the SMs with the reconstructed images.
In \cite{fabian2022humus}, convolutional blocks were used to extract high-resolution features and transformer blocks have been introduced to refine low-resolution features.
The authors of \cite{sun2023joint} used pre-acquired intra-subject MRI modalities to guide the reconstruction process.
In \cite{hu2022trans}, the authors designed a transformer-enhanced network, incorporating a regularization term on the error maps within the residual image domain.
Some works solved the optimization problem with structured low-rank algorithm \cite{pramanik2020deep, zhang2022accelerated, qiu2021automatic}.
The authors in \cite{wang2022one} designed a deep low-rank and sparse network which explored 1D convolution during training.
In \cite{yiasemis2023vsharp}, ADMM was utilized to unroll a half-quadratic variable splitting based optimization process, and a dilated-convolution model was designed to predict the Lagrange multipliers.
However, the complementarity between unrolled networks and distribution-learning-based methods has not been thoroughly investigated; in this study, we propose a novel concept in which we treat naive sampling results from SMLD methods as guidance and unrolled cascades as denoising and refining processes.

\section{Background}
\label{sec:background}
In PI, the acquisition of multicoil measurements can be formulated as
\begin{equation}\label{sampling}
	y = \mathcal{A}x+b
\end{equation}
where $y\in\mathbb{C}^{NC}$ is the undersampled k-space data and $C$ is the number of coils.
$x\in\mathbb{C}^{M}$ is the underlying MR image for reconstruction and $N<M$.
$b$ is the measurement noise.
$\mathcal{A}$ is the forward operator given by
\begin{equation}\label{forward}
	\mathcal{A}(\cdot)=\mathcal{P}\circ\mathcal{F}\circ\mathcal{E}(\cdot)
\end{equation}
where $\mathcal{P}$ denotes the sampling masks that zero the undersampled data points, and $\mathcal{F}$ denotes the Fourier transform matrix.
The expanding operator $\mathcal{E}(\cdot)$ calculates the coil-specific images from $x$, defined as
\begin{equation}\label{expanding}
	\mathcal{E}(x) = (\mathcal{S}_1x, \mathcal{S}_2x, ...\mathcal{S}_Cx)=(x_1, x_2, ..., x_C), i=1,2,...,C
\end{equation}
where $\mathcal{S}_i$ is the SM of the $i$th coil.
Conversely, the reduced operator $\mathcal{R}$ integrates the coil-specific images into a single image defined as follows:
\begin{equation}\label{reducing}
	\mathcal{R}(x_1, x_2, ..., x_C)=\sum_{i=1}^C \mathcal{S}_i^*x_i
\end{equation}
and the backward operator is expressed as
\begin{equation}\label{backward}
	\mathcal{A}^*(\cdot)=\mathcal{R}\circ\mathcal{F}^{-1}\circ\mathcal{P}(\cdot) 
\end{equation}
It is ill-posed to recover missing data directly from $y$. 
Hence, a regularization term $\lambda\mathcal U(x)$ is routinely added to narrow the solution space, which is
\begin{equation}\label{objective}
	\hat{x}=\mathop{\arg\min}\limits_{x}\|\mathcal P\mathcal F \mathcal{E} x-y\|_{2}^{2}+\lambda\mathcal U(x)
\end{equation}
For example, $\lambda \Psi(x)$ is a common regularization term that imposes the sparsity constraint in CS-MRI.
Then, gradient descent can be used to solve the problem iteratively as follows:
\begin{equation}\label{GD}
	x^{t+1} = x^t - \alpha^t (\mathcal{A}^*(\mathcal{A}x^t-y)+\lambda \psi(x^t))
\end{equation}
where $\alpha^t$ is the step size in the $t$th iteration and $\psi(x^t)$ is the gradient of $\Psi(x^t)$.

\section{Methods}
\label{sec:Method}
\begin{figure*}[!t]	
	\begin{minipage}[b]{1\linewidth}
		\centering
		\centerline{\includegraphics[width=14 cm]{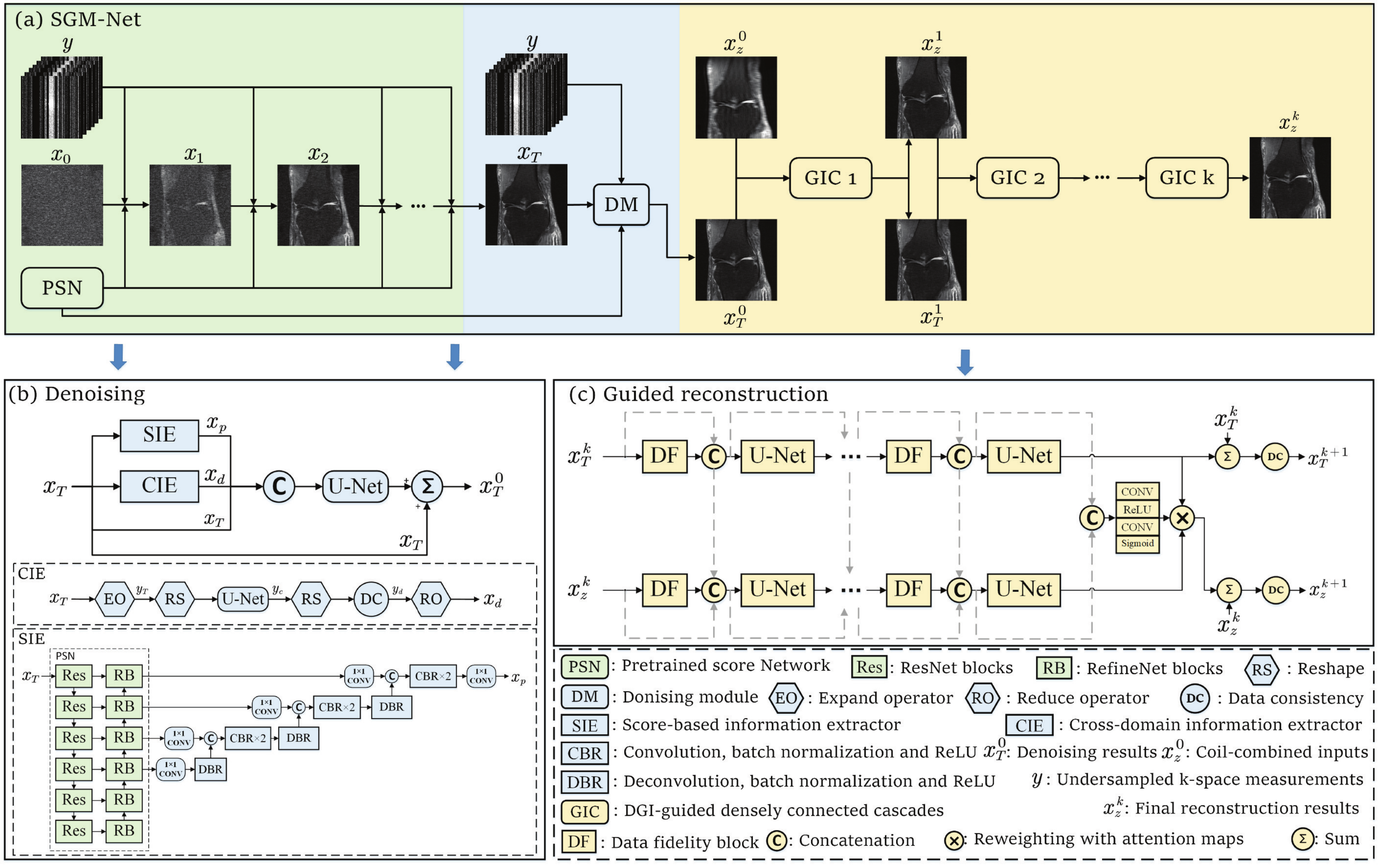}}
	\end{minipage}
	
	\caption{(a) The workflow of the proposed SGM-Net. (b) The network structure of the denoising module and the SIE and CIE blocks. (c) The network structure of the $(k+1)$th GIC. \label{fig:2}}
\end{figure*}
We proposed a score-based generative priors guided model-driven Network (SGM-Net), which is illustrated in Fig.~\ref{fig:2}.
The whole workflow is illustrated in Fig.~\ref{fig:2} (a), which comprises three parts: sampling (green components), denoising (blue components), and guided reconstruction (yellow components).
The preliminary samples (that is, $x_T$ in Fig.~\ref{fig:2} (a)) were first obtained using Langevin dynamics and then used as PGIs for the follow-up steps.
The gradients of the data distribution were obtained using a PSN.
Instead of carefully training and tuning the SMLD process to obtain high-quality samples, we sampled $x_T$ from an out-of-distribution pretrained network with fitted hyperparameters.
Then, we treated the artifacts in $x_T$ as removable and proposed a DM to enhance $x_T$.
We designed a SIE and a CIE to extract features in DMs, as shown in Fig.~\ref{fig:2} (b).
Finally, DGIs were used to guide an unrolled network training, as shown in Fig.~\ref{fig:2} (c).
Details of the proposed method are presented in the following subsections.

\subsection{Score matching and Langevin dynamics}
The first step of the proposed workflow is to obtain PGIs using SMLD method.
In this study, we adopted a PSN proposed in \cite{song2019generative}, which was successfully extended to MRI reconstruction in \cite{jalal2021robust}.
To improve the generality of our workflow, the score network was pretrained with out-of-distribution data, and the tedious tuning and retraining steps were bypassed.
The network, remarked as $s_{\theta}$, is a RefineNet \cite{lin2017refinenet} with dilated convolutions and conditional instance normalization, which is used to approach the score function of $p(x)$ (that is, $\bigtriangledown_x \log p(x)$) and $s_{\theta}\approx \bigtriangledown_x \log p(x)$.
$s_{\theta}$ is a six-layer U-shaped network.
The encoders are designed based on ResNet blocks and the decoders are based on RefineNet blocks.

The annealed Langevin method-based sampling process in MRI can be represented as
\begin{equation}\label{smld}
	x_{t+1} = x_t + \eta_t (s_\theta(x_t;\sigma_t) + \frac{\mathcal{A}^*(y-\mathcal{A}x_t)}{\gamma_t^2+\sigma^2}) + \sqrt{2\eta_t} \xi_t
\end{equation}
where $\xi_t\sim\mathcal{N}(0, \textit{I})$ and $t=0, 1, 2, .., T-1$. $\eta_t$ is the step size in the $t$th iteration and $\eta_t=\epsilon \cdot \frac{\sigma^2_j}{\sigma^2_L}$.
$\{\sigma_j\}_{j=1}^L$ is a sequence of noise scales that gradually reduces each step size in the annealed Langevin dynamics, where $\sigma_1 > \sigma_2 > \cdots > \sigma_L$.
$\frac{\mathcal{A}^*(y-\mathcal{A}x_t)}{\gamma_t^2+\sigma^2}$ is the consistency term introduced to exploit known measurements.
Finally, $x_T$ approaches the target as $T$ becomes sufficiently large, which is used as a guidance in the follow-up steps.

\subsection{Denoising module}
The PGIs generated in the first step exhibits over-smooth details and hallucination artifacts, and the denoising step aims to coarsely eliminate the noises and artifacts in PGIs.
To achieve this, we designed a CIE and a SIE block to extract features, which were then used within a residual structure to generate denoised outputs.
Consequently, the mappings between extracted features and the noise patterns were established.

The CIE blocks were designed to exploit the complementarity between the frequency domain and the image domain.
As depicted in Fig~\ref{fig:2} (b), first we used the expand operator to obtain the corresponding k-space data of $x_T$, marked as $y_T$.
Then, $y_T$ was reshaped and the coil channel was combined with the feature channel before fed to a U-Net.
In this study, all complex-valued MR images were stored in two distinct channels during the network training.
Therefore, the number of input and output channels of the U-Net was set to 30 (with 15-coil data).
A data consistency (DC) block was concatenated to the output of U-Net, marked as $y_c$, to maintain consistency with the measurements, and the $j$th coefficient in $y_c$ was updated as:
\begin{equation}\label{dc}
	y_d(j) = D(y_c(j))= \left\{ 
	\begin{array}{l}
		y_c(j), \ \ \ \ \ \ \ \ \ j\notin\Omega\\
		\frac{y_c(j)+\mu y(j)}{1+\mu}, \ \ \ \ j\in\Omega\\
	\end{array} \right.
\end{equation}
where $\Omega$ is the subset of sampled points from the complete k-space data.
$y_c$ and $y_d$ are the input and output of the DC blocks, respectively, and $y$ is the undersampled measurements.
We then used an reduce operator to transfer the k-space data back to the image domain, which is denoted as $x_{d}$.
The CIE blocks could extract the features associated with the noise patterns directly in k-space.
Considering the other components in DMs operate in the image domain, the output $x_{d}$ is expected to compensate for the missing information in the other blocks, thereby enhancing overall performance. 

Essentially, score network is used to generate clean images from noise, and the artifacts can be considered as generated during the sampling process.
Consequently, there could exist an underlying mapping between the artifacts and the elements involved in the score network.
Therefore, we designed a SIE to extract features correlated to the noise patterns from the PSN.
As depicted in Fig~\ref{fig:2} (b), the PSN is a six-layer RefineNet and the features in deeper RefineNet blocks (RBs) closely associated with the score of the inputs, which could potentially compromise noise-related information.
Moreover, considering that the optimal level of features for learning the noise is unrevealed, we fine-tuned the PSN and extract features across multiple layers.
Specifically, to build the mapping to the noise in $x_T$, we directly fed $x_T$ to PSN and during training, all the parameters in the PSN was fitted.
Then, we concatenated convolutional blocks to the last four RBs in PSN to learn multi-level features.
Each convolutions was followed by a batch normalization and a ReLU function.
Then, deconvolutions were used to double the feature sizes, and the features were concatenated with the counterparts extracted in the shallower RBs.
Each convolutions and deconvolutions was followed by a batch normalization block and a ReLU function.
$1\times1$ convolutions were first applied to the output of RBs to adjust the number of channels, thereby constraining the module scale.
As a results, different levels of features can be simultaneously captured to maximize the stability and accuracy of the mapping established to noise.

The output of the SIE is marked as $x_p$, which was concatenated with $x_d$ and fed to a U-Net, marked as $N_d$, to learn the noise in $x_T$.
Besides, to avoid newly introduced noises when obtaining $x_d$ and $x_p$ from $x_T$, $x_T$ was also fed to $N_d$.
Therefore, the enriched input features of $N_d$ have three components.
By concatenating and feeding them into the $N_d$, the denoising process is denoted as:
\begin{equation}
	x_T^0 = x_T + N_d\{x_T, x_p, x_{d}\}
\end{equation}
where $\{\cdot\}$ implements the channel-wise concatenation of its components.
With the skip connection of $x_T$, $N_d$ can directly learn the mapping to the noise from the enriched inputs.
Consequently, the noise and artifacts are expected to be further eliminated in $x_T^0$.

\subsection{Unrolled network with PGI-guided densely-connected cascades}
The DM aims to realize a coarse denoising, and the third step of the proposed workflow is to further improve the image quality
and recover the fine details from denoised guidance images (DGIs) (that is, $x_T^0$).
The undersampled measurements $x_z^0$ and $x_T^0$ were used as two different known information during unrolled network training.
As shown in Fig.~\ref{fig:2}, the network consists of several DGI-guided densely connected cascades (GICs).
The inputs of the $(k+1)$th GICs are $x_T^k$ and $x_z^k$ where $k=0, 1, ..., K-1$, and $x_z^0$ is the coil-combined zero-filled image of the undersampled measurements.
In each GIC, $x_z^k$ were periodically updated to approach the target images, whereas the DGIs $x_T^k$ were simultaneously updated to provide more accurate guidance, forming a parallel network structure.
$I$ Data fidelity (DF) and $I$ U-Net blocks were alternately stacked in each GIC to solve (7), and U-Net blocks served as learning based regularizers.
For the $x_T^k$ branch, the calculations in DF blocks can be represented as
\begin{equation}
	r_T^{k_{i+1}} = x_T^{k_i} - \alpha^{k_i} (\mathcal{A}^*(\mathcal{A}x_T^{{k_i}}-y)), i = 0, 1, ..., I-1
\end{equation}
where $x_T^{k_0}$ is equal to $x_T^{k}$. For the $x_z^k$-branch, the DF blocks are given by
\begin{equation}
	r_z^{k_{i+1}} = x_z^{k_i} - \alpha^{k_i} (\mathcal{A}^*(\mathcal{A}x_z^{k_i}-y)), i = 0, 1, ..., I-1
\end{equation}
where $x_z^{k_0}$ is equal to $x_z^{k}$.
In addition, we proposed a densely connected structure to enrich the input information for each U-Net block.
This is based on the idea that DL-based methods are more flexible than the conventional algorithms, allowing for learning from richer inputs across various stages and extracting more valuable features.
Consequently, the densely connected information can mitigate the impact of inaccurate cascades and establish robust mappings to the targets.
For the $x_T^k$-branch, the initial input and each output of DF blocks corresponding to a U-Net block were concatenated before being fed into U-Net.
Therefore, we have:
\begin{equation}
	x_T^{k_{i+1}} = U_T^{i+1}\{x_T^k, r_T^{k_1}, r_T^{k_2}, ..., r_T^{k_{i+1}}\}, i = 0, 1, ..., I-1
\end{equation}
where $U_T^{i+1}$ denotes the $(i+1)$th U-Net block in the $x_T^k$-branch, $x_T^k$ denotes the input to the corresponding branch, and $r_T^{k_i}$ denotes the output of the $i$th DF block.
A similar strategy was implemented in the $x_z^k$-branch.
Considering that we take the final output of the $x_z^k$ branch in the last GIC as the final reconstruction, the $x_z^k$ branch is essentially the main branch, and the $x_T^k$ branch is simultaneously used to guide the reconstruction in the main branch.
Therefore, the concatenated information from the $x_T^k$ branch was also fed to the $x_z^k$ branch to learn from the features in different stages, represented as
\begin{equation}
	x_z^{k_{i+1}} = U_z^{i+1}\{x_T^k, r_T^{k_1}, r_T^{k_2}, ..., r_T^{k_{i+1}}, x_z^k, r_z^{k_1}, r_z^{k_2}, ..., r_z^{k_{i+1}}\}, i = 0, 1, ..., I-1
\end{equation}

After the DF and U-Net blocks, we designed a skip connection and a DC block at the end of the $x_T^k$ branch, and the produced $x_T^{k+1}$ is fed to the next GIC as one of the two inputs as follows:
\begin{equation}
	x_T^{k+1} = D(x_T^{k} +x_T^{k_{I}})
\end{equation}
For $x_z^k$, we further incorporated a four-layer attention module (AM), marked as $\mathcal{M}_A$, between the last U-Net and skip connection, as shown in Fig~\ref{fig:2} (c).
Information fed to the last U-Nets in the two branches were concatenated and simultaneously fed to the AM to produce an attention map
\begin{equation}
	m_A = \mathcal{M}_A\{x_T^k, r_T^{k_1}, r_T^{k_2}, ..., r_T^{k_{I}}, x_z^k, r_z^{k_1}, r_z^{k_2}, ..., r_z^{k_{I}}\}
\end{equation}
Through a skip connection and DC block, $x_z^{k+1}$ is given as
\begin{equation}
	x_z^{k+1} = D(x_z^{k} + x_z^{k_I} \odot m_A + x_T^{k_{I}} \odot ((\textbf{J} - m_A))
\end{equation}
where $\textbf{J}$ is an all-ones matrix and $\odot$ denotes the Hadamard product. The element-wise attention results in a deeper fusion between the two branches and optimizes the intermediate results in each $x_z^k$ branch, which amplifies the likelihood for subsequent GICs to converge towards the targets.

\subsection{loss functions}
In this study, the overall loss for network training was designed as:
\begin{equation}
	\mathcal{L}{(x_T^0, x_g)} + \mathcal{L}{(x_z^K, x_g)} +\sum_{k=1}^K w_k [ \mathcal{L}{(x_T^{k}, x_g)} + \mathcal{L}{(x_z^{k}, x_g)} ]
\end{equation}
where $x_g$ is the target image. $x_T^0$ amd $x_z^K$ are the final outputs of DM and the unrolled network. $x_T^{k}$ and $x_z^{k}$ are intermediate results from the two branches in the unrolled network. 
$\{w_k\}_{k=1}^K$ is a sequence assigns heavier weights to the intermediate results from deeper GICs and $w_k=10^{\frac{k-K}{K-1}}$ \cite{yiasemis2023vsharp}.
When training on the coronal PD sequence and FastMRI dataset, we adopted mean square error as the loss function that $\mathcal{L}:=\mathcal{L}_{MSE}$.
For the coronal PDFS sequence, we further included a SSIM loss that $\mathcal{L}:=\mathcal{L}_{MSE} + \mathcal{L}_{SSIM}$ which equally contribute with the MSE loss during training to improve reconstruction outcomes.

\subsection{Implementation details}
We adopted the score network $s_{\theta}$ released by the authors in \cite{jalal2021robust}, which was pretrained on FastMRI brain dataset. 
We followed the proposed hyperparameters in that manuscript\footnote{https://github.com/utcsilab/csgm-mri-langevin} during sampling, and we did not retrain the network on our knee dataset.
In SGM-Net, $K$ was set to five and $I$ was set to three. The numbers of filters in the U-Nets from DMs were set to 32, 64, 128, and 256. The numbers of filters in all U-Nets from GICs were set to 16, 32, 64, and 128, respectively.
In SIE blocks, the number of output channels of the RBs were set to 128, 128, 256, 256, 256 and 512.
The number of channels of the inputs and outputs in $1\times1$ convolutions, which are concatenated to RBs, were set to (128, 16), (128, 32), (256, 64) and (256, 128).
For the last $1\times1$ convolution, the number of input and output channels were 16 and 2, respectively.
The CBRs and DBRs halved the number of channels of their inputs.
The step size in DF blocks and the $\mu$s in DC blocks were learned directly in the network.
The proposed SGM-Net was trained on a UBUNTU 20.04 LTS system with a 3090 GPU (24GB) for 100 epoches, and the learning rate was set to 1e-3. The batch size was set to 1.
\section{Experiments and Results}
\label{sec:ear}

\subsection{Dataset}

\begin{figure}[!t]	
	\begin{minipage}[b]{1\linewidth}
		\centering
		\centerline{\includegraphics[width=14cm]{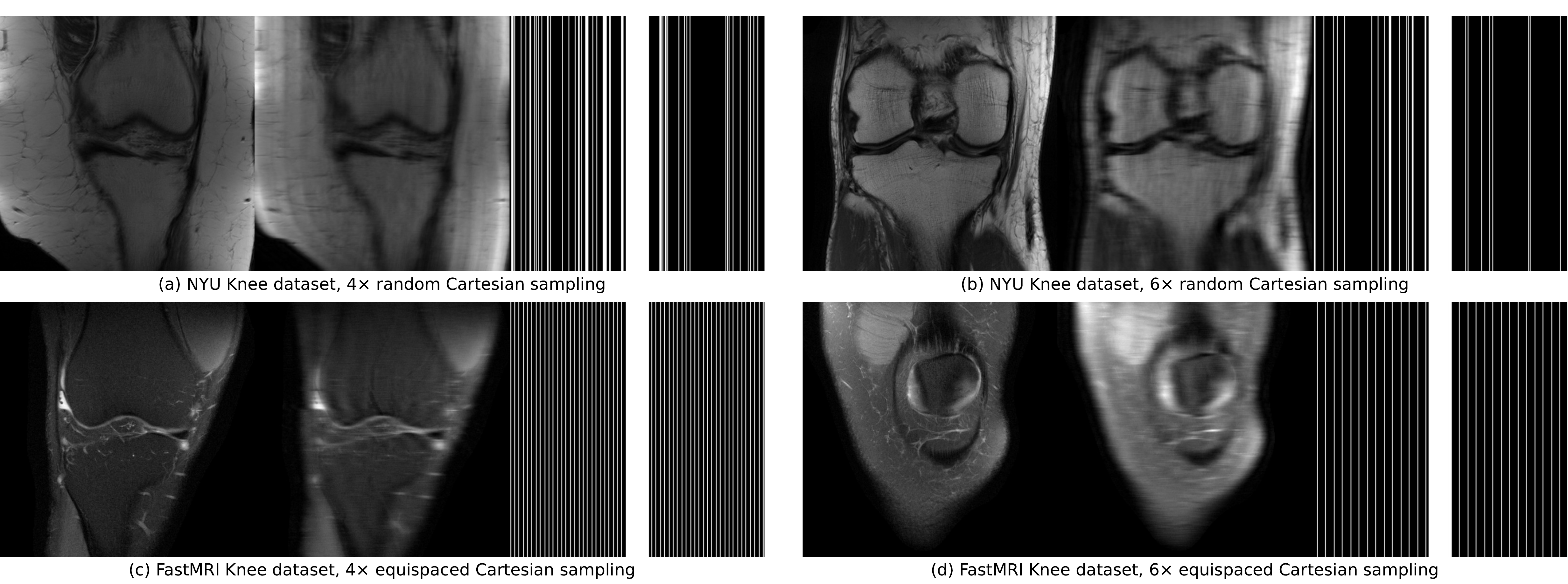}}
	\end{minipage}
	
	\caption{Examples of ground truth images (the first column), undersampled images (the second column) and the corresponding k-space trajectories (the third column) from different datasets.\label{fig:3}}
\end{figure}

In this study, we adopted a publicly available 15-coil NYU knee dataset \cite{hammernik2018learning} to train our model and compared it with other methods.
The knee dataset consisted of five 2D turbo spin-echo (TSE) sequences, each of which contained volumes scanned using a 3T scanner from 20 patients.
Each volume consists of approximately 40 slices. 
We conducted experiments on coronal proton density (PD; TR=2750 ms, TE=27 ms, in-plane resolution 0.49$\times$0.44 mm$^2$, slice thickness 3 mm, 35-42 slices), and coronal fat-saturated PD (PDFS; TR=2870 ms, TE=33 ms, in-plane resolution 0.49$\times$0.44 mm$^2$, slice thickness 3 mm, 33-44 slices) sequences.
The coil sensitivity maps were precomputed using ESPRIRiT and included in the dataset. 
The training, validation, and testing sets were randomly divided at a ratio of 3:1:1.
Random Cartesian Sampling masks are applied to obtain undersampled measurements with four-fold and six-fold acceleration. 8\% lines at the center region are preserved. 

FastMRI \cite{zbontar2018fastmri} is a publicly available dataset for accelerated MRI, which has widely been adopted in relevant studies.
Considering the limitations of our hardware, we randomly selected a subset of the 15-coil knee dataset to evaluate our model and demonstrate its effectiveness.
The data were acquired using the 2D TSE protocol, containing PD and PDFS sequence scanning from 3T or 1.5T systems (TR=2200-3000 ms, TE=27-34 ms, in-plane resolution 0.5$\times$0.5 mm$^2$, slice thickness 3 mm).
A total of 419, 33, and 375 slices were used for training, validation, and testing sets, respectively.
Coil sensitivity maps are not available in the dataset, and related studies have been conducted to estimate and optimize sensitivity maps during training.
However, we did not focus on SM estimation in this study and used pretrained SMs in our model to highlight the improvements brought about by the proposed methods.
In the FastMRI dataset, we calculated the SMs using the sigpy \cite{martin2020sigpy} toolbox.
Equispaced Cartesian Sampling masks are applied to obtain undersampled measurements with four-fold and six-fold acceleration. 8\% lines at the center region are preserved. Examples of training images and the corresponding k-space sampling trajectories are shown in Fig.~\ref{fig:3}.

\subsection{comparisons with the cutting-edge methods}

\begin{table*}[!t]
	\centering
	\caption{The quantitative results on Coronal PD and PDFS sequences, as well as the FastMRI dataset, compared with those of cutting-edge methods.}\label{sota}
	\resizebox{1.\columnwidth}{!}
	{\begin{tabular}{cccccc} 
			\toprule
			\multirow{2}{*}{Sequence} & \multirow{2}{*}{Method} & \multicolumn{2}{c}{PSNR} & \multicolumn{2}{c}{SSIM} \\
			& & 4× & 6× & 4× & 6×\\
			\hline 
			\multirow{11}{*}{Coronal PD} & zero-filled & 31.3089$\pm$3.3939 & 30.5176$\pm$3.3902 & 0.8778$\pm$0.0713 & 0.8626$\pm$0.0799\\
			& TV & 33.7498$\pm$3.1753 & 32.3290$\pm$3.2159 & 0.8851$\pm$0.0657 & 0.8658$\pm$0.0725\\
			& U-Net & 36.8984$\pm$2.7072 & 34.7352$\pm$2.7863 & 0.9361$\pm$0.0610 & 0.9133$\pm$0.0682\\
			& D5C5 & 38.7903$\pm$2.8408 & 35.9697$\pm$2.7990 & 0.9509$\pm$0.0587 & 0.9259$\pm$0.0689 \\
			& ISTA-Net & 39.2910$\pm$2.8423 & 35.5220$\pm$3.0915 & 0.9471$\pm$0.0567 & 0.9138$\pm$0.0717\\
			& VS-Net & 40.1649$\pm$2.8767 & 37.0604$\pm$2.7927 & 0.9591$\pm$0.0569 & 0.9358$\pm$0.0648\\
			& E2E-VarNet & 39.3592$\pm$2.9368 & 36.8949$\pm$3.3872 & 0.9542$\pm$0.0662 & 0.9342$\pm$0.0775\\
			& ReVarNet & 40.2545$\pm$2.8766 & 37.3683$\pm$2.7869 & 0.9590$\pm$0.0588 & 0.9379$\pm$0.0677\\
			& MeDL-Net & 40.9295$\pm$2.9072 & 38.1052$\pm$2.7622 & 0.9614$\pm$0.0557 & 0.9417$\pm$0.0663\\
			& vSharp & 40.3121$\pm$2.7773 & 37.5657$\pm$2.5346 & 0.9341$\pm$0.0575 & 0.8982$\pm$0.0660\\
			& SGM-Net (ours) & \textbf{41.1981$\pm$3.3802} & \textbf{38.4491$\pm$3.2166} & \textbf{0.9634$\pm$0.0682} & \textbf{0.9452$\pm$0.0790}\\
			\hline 
			\multirowcell{11}{Coronal PDFS} & zero-filled & 32.0299$\pm$2.0273 & 31.2221$\pm$2.0575 & 0.7964$\pm$0.0998 & 0.7587$\pm$0.1172\\
			& TV & 33.7256$\pm$1.9058 & 32.7328$\pm$1.8850 & 0.8124$\pm$0.0752 & 0.7920$\pm$0.0796\\
			& U-Net & 34.4367$\pm$0.6572 & 33.2028$\pm$2.5662 & 0.8183$\pm$0.0106 & 0.7758$\pm$0.1222\\
			& D5C5& 34.9671$\pm$2.8261 & 33.4189$\pm$2.4765 & 0.8250$\pm$0.1106 & 0.7854$\pm$0.1223 \\
			& ISTA-Net & 34.4990$\pm$2.6389 & 33.4569$\pm$2.5758 & 0.8198$\pm$0.1066 & 0.7854$\pm$0.1214\\
			& VS-Net & 34.6847$\pm$2.7134 & 33.5010$\pm$2.6100 & 0.8259$\pm$0.1073 & 0.7842$\pm$0.1229\\
			& E2E-VarNet & 34.6274$\pm$2.7331 & 33.1626$\pm$2.4742 & 0.8234$\pm$0.1101 & 0.7804$\pm$0.1244 \\
			& ReVarNet & 34.8522$\pm$2.7461 & 33.3146$\pm$2.5604 & 0.8272$\pm$0.1080 & 0.7832$\pm$0.1232\\
			& MeDL-Net & 35.5931$\pm$2.7307 & 34.1409$\pm$2.5501 & 0.8349$\pm$0.1025 & 0.7922$\pm$0.1255\\
			& vSharp & 35.5536$\pm$2.6244 & 34.3442$\pm$2.4485 & 0.8495$\pm$0.0940 & 0.8210$\pm$0.0990\\
			& SGM-Net (ours) & \textbf{36.1426$\pm$2.6821} & \textbf{34.7450$\pm$2.6737} & \textbf{0.8648$\pm$0.0946} & \textbf{0.8395$\pm$0.0987}\\
			\hline
			\multirow{11}{*}{FastMRI} & zero-filled & 33.2828$\pm$3.0594 & 32.5069$\pm$3.0657 & 0.8857$\pm$0.0709 & 0.8680$\pm$0.0791\\
			& TV & 35.5512$\pm$2.9189 & 33.8441$\pm$2.8703 & 0.8884$\pm$0.0635 & 0.0201$\pm$0.0152\\
			& U-Net & 35.2158$\pm$2.5925 & 33.7864$\pm$2.5331 & 0.8562$\pm$0.0574 & 0.8091$\pm$0.0637\\
			& D5C5 & 36.4550$\pm$3.0246 & 34.5636$\pm$2.8510 & 0.9113$\pm$0.0649 & 0.8849$\pm$0.0749\\
			& ISTA-Net & 36.5753$\pm$2.9887 & 34.6229$\pm$2.8269 & 0.9134$\pm$0.0656 & 0.8837$\pm$0.0752\\
			& VS-Net & 37.3042$\pm$3.1130 & 35.5335$\pm$2.8513 & 0.9202$\pm$0.0637 & 0.8916$\pm$0.0736\\
			& E2E-VarNet & 37.5951$\pm$3.0129 & 35.6696$\pm$2.8326 & 0.9217$\pm$0.0635 & 0.8978$\pm$0.0726\\
			& ReVarNet & 37.7652$\pm$3.0632 & 35.7778$\pm$2.6971 & 0.9295$\pm$0.0589 & 0.9084$\pm$0.0675\\
			& MeDL-Net & 37.6791$\pm$2.9644 & 35.9594$\pm$2.7401 & 0.8984$\pm$0.0643 & 0.8882$\pm$0.0717\\
			& vSHARP & 38.0836$\pm$3.1283 & 35.7326$\pm$2.8239 & 0.9134$\pm$0.0645 & 0.8621$\pm$0.0917\\
			& SGM-Net (ours)  & \textbf{38.8241$\pm$3.3808} & \textbf{36.8730$\pm$3.0364} & \textbf{0.9344$\pm$0.0588} & \textbf{0.9094$\pm$0.0707}\\
			\bottomrule
		\end{tabular}
	}
\end{table*}

\begin{figure*}[!t]	
	\begin{minipage}[b]{1\linewidth}
		\centering
		\centerline{\includegraphics[width=13.5 cm]{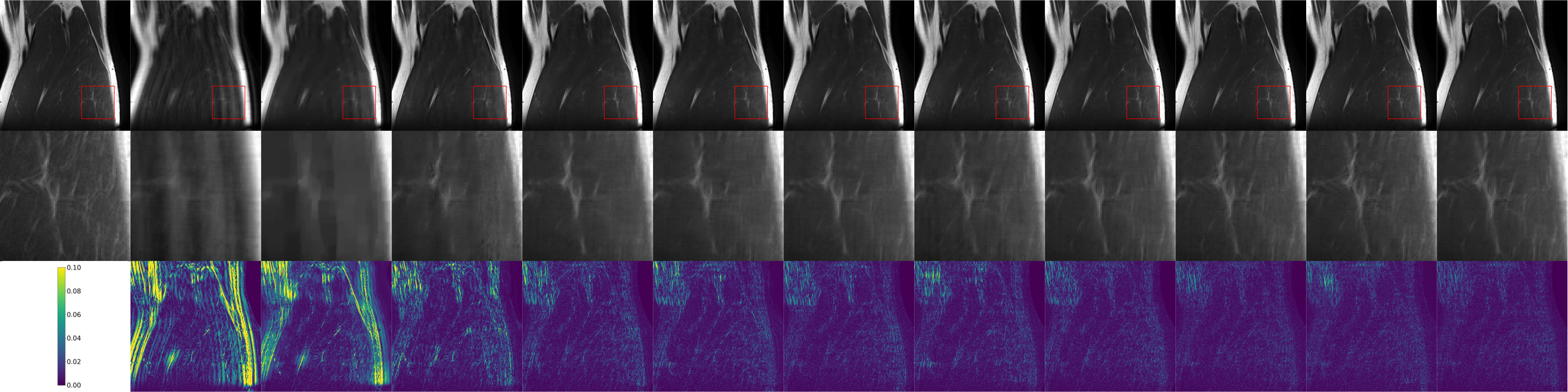}}
		\begin{center}
			\vspace{-0.1 cm}
			\footnotesize coronal PD 4$\times$ acceleration
			\vspace{0.01 cm}
		\end{center}
		
	\end{minipage}
	\begin{minipage}[b]{1\linewidth}
		\centering
		\centerline{\includegraphics[width=13.5 cm]{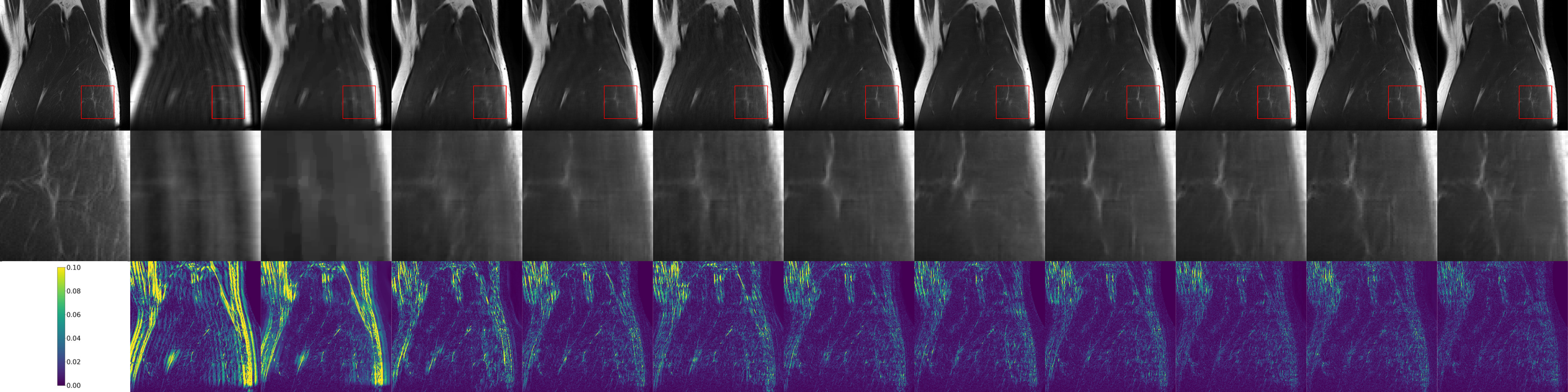}}
		\begin{center}
			\vspace{-0.1 cm}
			\footnotesize coronal PD 6$\times$ acceleration
		\end{center}
		
	\end{minipage}
	\begin{minipage}[b]{1\linewidth}
		\centering
		\centerline{\includegraphics[width=13.5 cm]{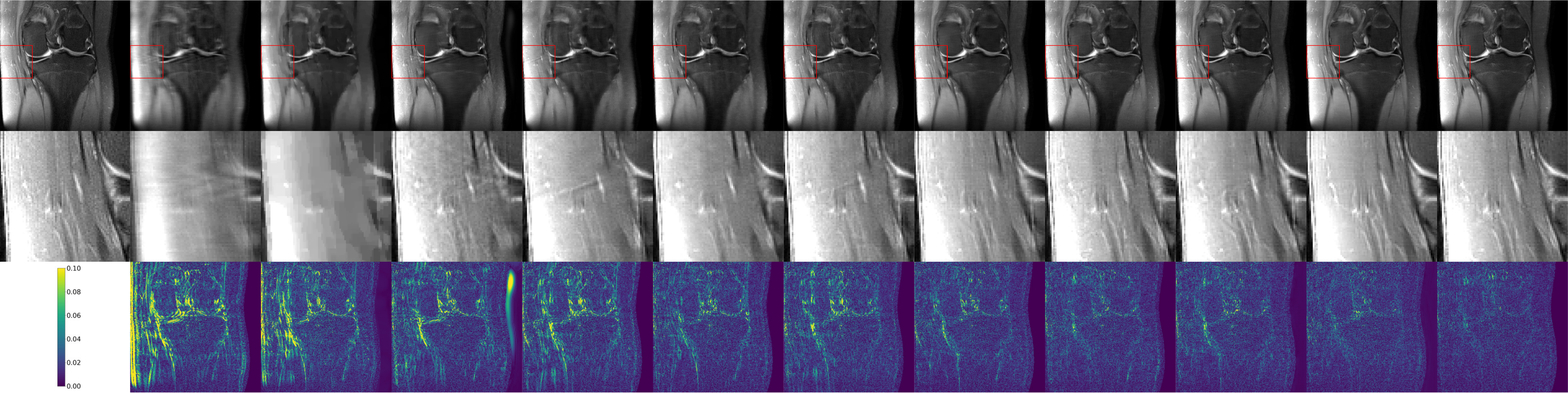}}
		\begin{center}
			\vspace{-0.1 cm}
			\footnotesize FastMRI 4$\times$ acceleration
		\end{center}
		
	\end{minipage}
	\begin{minipage}[b]{1\linewidth}
		\centering
		\centerline{\includegraphics[width=13.5 cm]{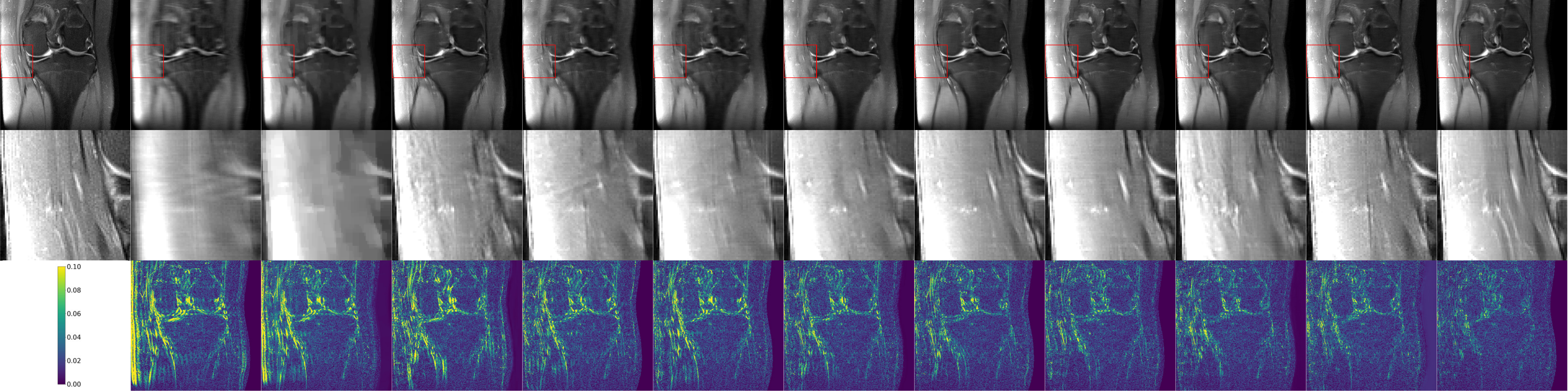}}
		\begin{center}
			\vspace{-0.1 cm}
			\footnotesize FastMRI 6$\times$ acceleration
		\end{center}
		
	\end{minipage}
	\caption{Examples of the reconstructed images of different methods from coronal PD sequence and FastMRI dataset at 4$\times$ and 6$\times$ acceleration. The first rows of each subfigure are the reconstructed images, the second rows are zoomed details in the red square, and the third rows are the error maps corresponding to the ground truth images. The methods from left to right: 1. ground truth; 2. zero-filled; 3. TV; 4. U-Net; 5. D5C5; 6. ISTA-Net; 7. VS-Net; 8. E2EVarNet; 9. ReVarNet; 10. MeDL-Net; 11. vSHARP; 12. SGM-Net (ours). \label{fig:4}}
\end{figure*}
We compared the proposed SGM-Net with two traditional and eight DL-based methods, as presented in Table \ref{sota}.
In the zero-fill method, the missing data in the measurements were filled with zero.
The TV method used total variation as the image prior in a CS-based approach.
For DL-based methods (\textit{i.e.}, U-Net \cite{zbontar2018fastmri}\footnote{https://github.com/facebookresearch/fastMRI},
D5C5 \cite{schlemper2017deep}\footnote{https://github.com/js3611/Deep-MRI-Reconstruction}. 
ISTA-Net \cite{zhang2018ista}\footnote{https://github.com/jianzhangcs/ISTA-Net-PyTorch}
VSNet \cite{duan2019vs}\footnote{https://github.com/j-duan/VS-Net}.
E2EVarNet \cite{sriram2020end}\footnote{https://github.com/NKI-AI/direct\label{direct}},  
ReVarNet \cite{yiasemis2021recurrent}\textsuperscript{\ref{direct}}, MeDL-Net \cite{qiao2023medl}\footnote{https://github.com/joexy312/MEDL-Net} and vSHARP \cite{yiasemis2023vsharp}\textsuperscript{\ref{direct}}), we adopted the released codes and retrained the networks on our device.
We made a concerted effort to keep the announced settings same as those in the corresponding articles, except that we reduced the number of filters in ReVarNet so that all networks could be trained on one GPU.
All methods were implemented on the knee dataset (coronal PD and PDFS sequences) and a subset of fastMRI dataset.
The PSNR and SSIM results were calculated under four- and six-fold acceleration to evaluate different methods.
As shown in Table \ref{sota}, the proposed SGM-Net robustly achieves the best results on all the selected datasets with different acceleration factors. 
Examples of reconstruction images are presented in Fig.\ref{fig:4}.
We can observe from the zoomed region (second rows) that our method recovered the most details, and the error maps (third row) exhibited the least errors.

\subsection{Ablation Studies}

\begin{table*}[!t]
	\caption{Results of ablation studies on coronal PD sequence with four- and six-fold acceleration rates.}\label{tab:ablation}
	\centering
	\resizebox{1.\columnwidth}{!}{
		\begin{tabular}{ccccc} 
			\toprule
			\multirow{2}{*}{Method}  & \multicolumn{2}{c}{PSNR} & \multicolumn{2}{c}{SSIM} \\
			& 4× & 6× & 4× & 6× \\
			\hline 
			SG & 31.3478$\pm$2.6593 & 29.7293$\pm$5.1889 & 0.4730$\pm$0.1556 & 0.6067$\pm$0.1215 \\
			MN  & 38.2080$\pm$2.4697 & 35.8126$\pm$2.4974 & 0.8346$\pm$0.1386 & 0.7849$\pm$0.1279\\
			SG-MN & 35.4585$\pm$2.9562 & 33.2093$\pm$2.6398 & 0.8351$\pm$0.0912 & 0.7412$\pm$0.1284\\	
			SG-MN-DM  & 39.3722$\pm$3.6295 & 36.9837$\pm$4.0466 & 0.9533$\pm$0.0801 & 0.9368$\pm$0.0812\\
			SG-MN-DM-US  & 40.5941$\pm$3.0464 & 37.8509$\pm$2.8824 & 0.9615$\pm$0.0619 & 0.9425$\pm$0.0684\\
			SG-MN-DM-DG  & 41.0969$\pm$3.7561 & 38.3457$\pm$2.9926 & 0.9625$\pm$0.0803 & 0.9440$\pm$0.0730\\
			SG-MN-US-DG  & 41.1213$\pm$3.3004 & 38.3419$\pm$3.0769 & 0.9630$\pm$0.0715 & 0.9444$\pm$0.0772\\
			SG-MN-DM-US-DG  & \textbf{41.1981$\pm$3.3802} & \textbf{38.4491$\pm$3.2166} & \textbf{0.9634$\pm$0.0682} & \textbf{0.9452$\pm$0.0790}\\
			\bottomrule 
	\end{tabular}}
\end{table*}

\begin{table*}[!t]
	\caption{Reconstruction results on coronal PD sequence with four- and six-fold acceleration rates using different inputs in DMs.}\label{tab:ablation2}
	\centering
	\resizebox{1.\columnwidth}{!}{
		\begin{tabular}{cccccc} 
			\toprule
			\multirow{2}{*}{structure} &\multirow{2}{*}{Method}  & \multicolumn{2}{c}{PSNR} & \multicolumn{2}{c}{SSIM} \\
			&	& 4× & 6× & 4× & 6× \\
			\hline 
			\multirow{4}{*}{w/o unrolled network} & $x_T$ & 38.5337$\pm$3.8373 & 36.2220$\pm$4.5657 & 0.9487$\pm$0.0812 & 0.9322$\pm$0.0825\\
			& $x_T, x_d$ & 38.9168$\pm$3.9265 & 36.4447$\pm$4.5595  & 0.9496$\pm$0.0808 & 0.9327$\pm$0.0818\\
			& $x_T, x_p$ & 38.9123$\pm$3.4768 & 36.6218$\pm$4.1896 & 0.9507$\pm$0.0799 & 0.9345$\pm$0.0814\\
			& $x_T, x_d, x_p$ & \textbf{39.3722$\pm$3.6295} & \textbf{36.9837$\pm$4.0466} & \textbf{0.9533$\pm$0.0801} & \textbf{0.9368$\pm$0.0812}\\
			\hline 
			
			\multirow{4}{*}{w unrolled network} & $x_T$ & 41.1164$\pm$3.4041 & 38.2711$\pm$3.6271& 0.9629$\pm$0.0708 & 0.9441$\pm$0.0825\\
			& $x_T, x_d$ & 41.0953$\pm$3.3920 & 38.2932$\pm$3.3948 & 0.9628$\pm$0.0669 & 0.9438$\pm$0.0814\\
			& $x_T, x_p$ & 41.1620$\pm$3.1978 & 38.4075$\pm$3.3948 & 0.9633$\pm$0.0639 & 0.9450$\pm$0.0812\\
			& $x_T, x_d, x_p$ & \textbf{41.1981$\pm$3.3802} & \textbf{38.4491$\pm$3.2166} & \textbf{0.9634$\pm$0.0682} & \textbf{0.9452$\pm$0.0790}\\
			\bottomrule 
	\end{tabular}}
\end{table*}

To highlight the improvements brought by different modules, we designed several ablation workflows for a clear evaluation.
Specifically, the proposed modules for evaluation were the SMLD-based guidance (SG), model-driven network structure (MN), denoising module (DM), updating steps (US) for SGs, and densely connected guidance (DG) in the unrolled network.
The ablations was designed as follows.
\begin{enumerate}
	\item SG: using the SMLD results as the final reconstructions.
	\item MN: a standard unrolled network without guidances. The undersampled measurements are adopted as the inputs. 
	\item SG-MN: a standard unrolled network with SMLD results as the inputs. 
	\item SG-MN-DM: The DGIs in SGM-Net are adopted as the final reconstructions.
	\item SG-MN-DM-US: the densely connected structure is removed from all the cascades.
	\item SG-MN-DM-DG: the SMLD results are used as guidance in the unrolled network but without the updating steps.
	\item SG-MN-US-DG: the SMLD results are used to guide the network without denoising by the DM.
	\item SG-MN-DM-US-DG: the proposed SGM-Net.
\end{enumerate}

The quantitative results are shown in table \ref{tab:ablation}.
Comparing SG with the proposed SGM-Net, we observed a significant improvement brought by the proposed modules, whereas the average results from the SMLD sampling were even worse than those of the zero-filled method.
Similarly, through the guidance of the SMLD results, the results of the unrolled MN method were significantly improved.
Using either of these two methods alone could not achieve satisfactory results.
Moreover, we simply combined SG and MN in the SG-MN workflow that the SMLD results were directly fed to the network as input.
However, the performance was worse than that of MN.
This phenomenon demonstrated the effectiveness of the proposed guidance-based workflow, which outperformed a simple combination of the two methods.

Comparing with the proposed SGM-Net, the SG-MN-DM workflow removed the guided reconstruction structures, and we observed deterioration on the results.
On the other hand, the improvements brought by the DMs over SG and MN method were also obvious.

A comparison between SG-MN-DM-US and SGM-Net demonstrated that the densely connected structure was capable of fulfilling the guiding work and significantly improving the final results.
Removing the densely connected structure and using only the AMs to optimize the reconstructed images would lead to a degredation in the final results.
By comparing SG-MN-DM-DG with SGM-Net, we observed that the updating steps in the proposed model can further ensured that accurate guidance was provided by the PGIs and improved the results.
When removing the DM and directly feeding SMLD results to guide the network, the reconstruction performances of SG-MN-US-DG were similarly worse than that of SGM-Net.

Meanwhile, when removing DM or US structure from SGM-Net, we observed that the network still achieved convincing results.
Specifically, when removing US, the denoised output of DM can provide effective guidance for the subsequent reconstructions.
When removing DM, the updating steps in the networks improved the image quality of SMLD results during training.
This phenomenon further demonstrated the robustness of the proposed network structure.
The best results were achieved with the integration of all the proposed modules, which demonstrated their effectiveness and necessity.
Furthermore, we illustrate in subsection~\ref{ltd} that the integration of the proposed module can enhance the robustness of the network with greatly reduced training data.

Besides, we conducted experiments using different features in DMs, and the results are shown in Table \ref{tab:ablation2}.
Here, we evaluated the results with and without the guided reconstruction process.
When the unrolled network structure was removed, the improvements brought by involving $x_d$ or $x_p$ in DMs is evident, which clearly exhibited the improvements brought by the SIE and CIE blocks.
With the unrolled network structure, the distances among the methods were narrowed, that the unrolled network can compensate for the deterioration in DMs.
However, incorporating all the components in DMs exhibited the best results, demonstrating the necessity of the proposed SIE and CIE blocks.

\begin{figure*}[!t]	
	\begin{minipage}[b]{0.49\linewidth}
		\centering
		\centerline{\includegraphics[width=\textwidth]{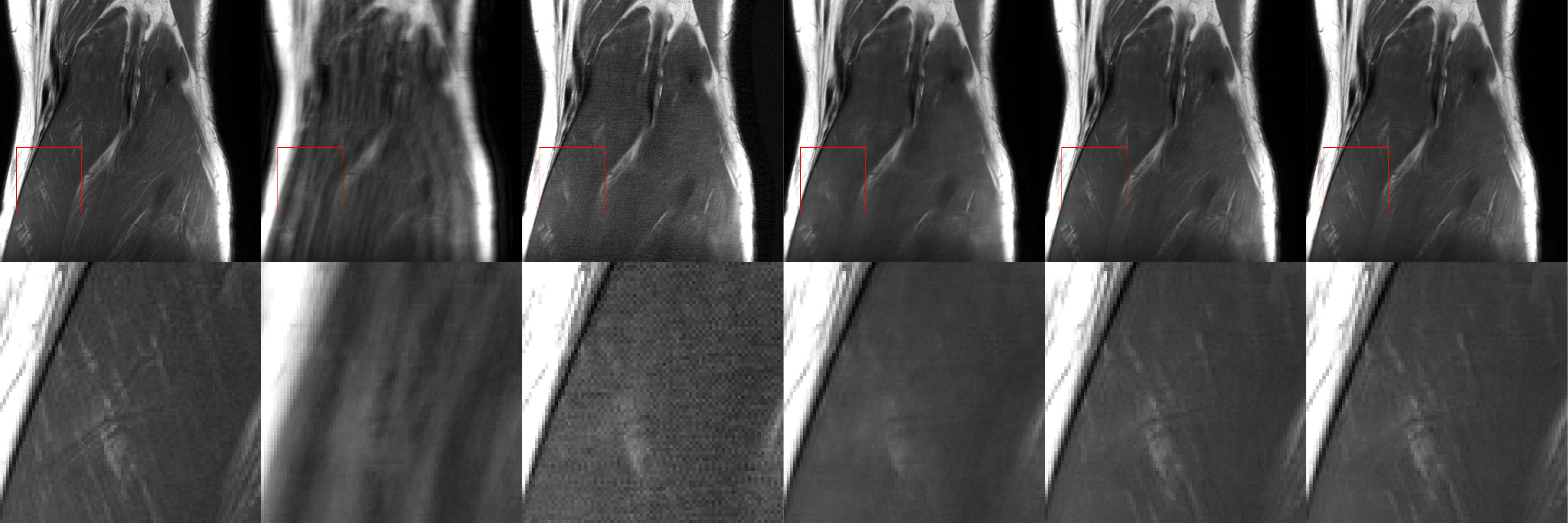}}
		\footnotesize (a)
	\end{minipage}
	\begin{minipage}[b]{0.49\linewidth}
		\centering
		\centerline{\includegraphics[width=\textwidth]{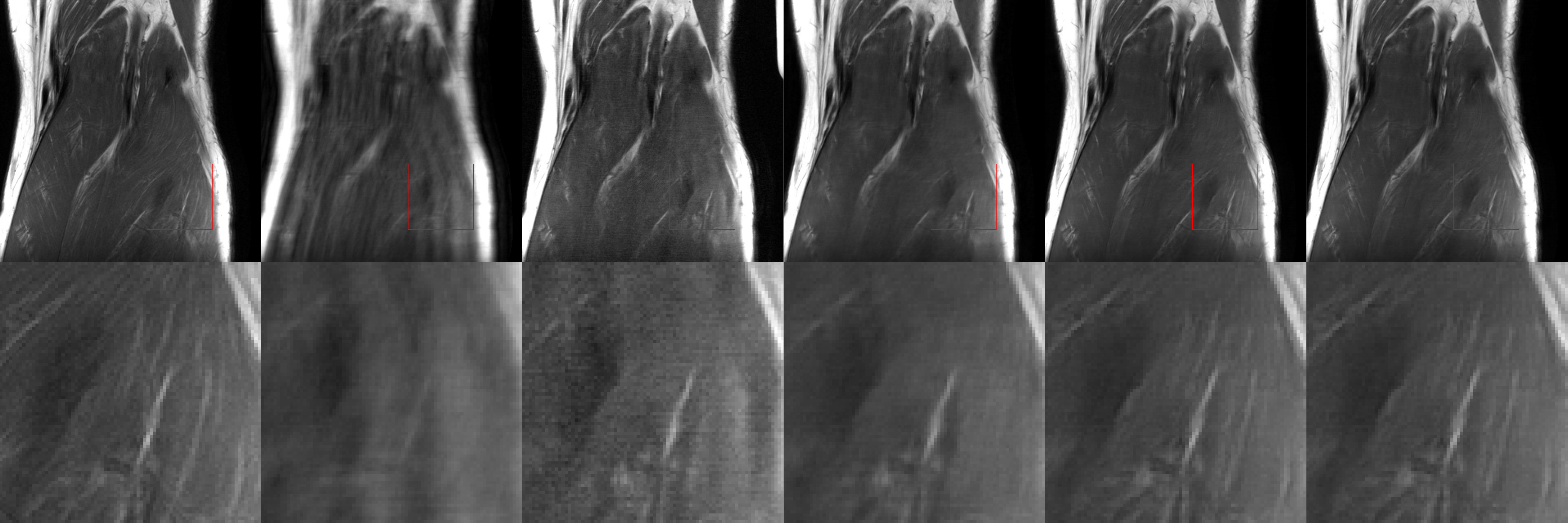}}
		\footnotesize (b)
	\end{minipage}
	\caption{(a) The reconstruction results of the original model at four-fold acceleration. (b) The reconstruction results with 20\% sampling steps at four-fold acceleration. The first rows of each subfigure are the reconstructed images, and the second rows are zoomed details in the red squares. The corresponding images from left to right: the ground truth, zero-filled measurements, PGI, DGI, and the final output of $x_T^K$ and $x_z^K$ branch. \label{fig:5}}
\end{figure*}

\begin{figure*}[!t]	
	\begin{minipage}[b]{0.49\linewidth}
		\centering
		\centerline{\includegraphics[width=\textwidth]{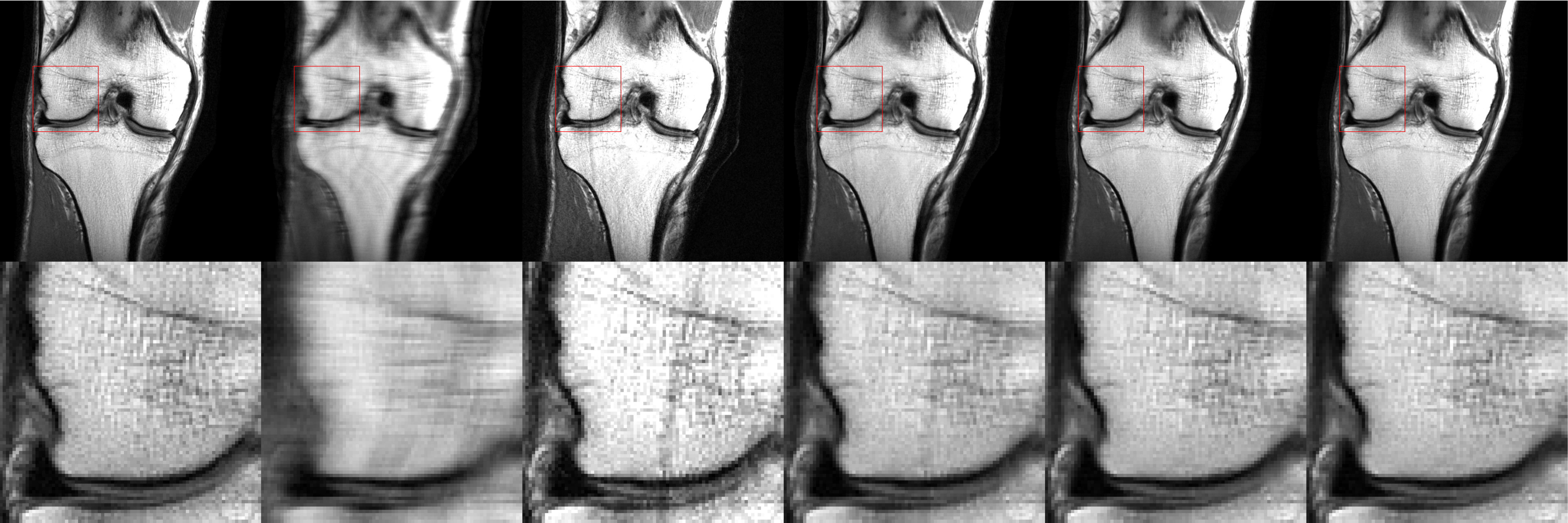}}
		\footnotesize (a)
	\end{minipage}
	\begin{minipage}[b]{0.49\linewidth}
		\centering
		\centerline{\includegraphics[width=\textwidth]{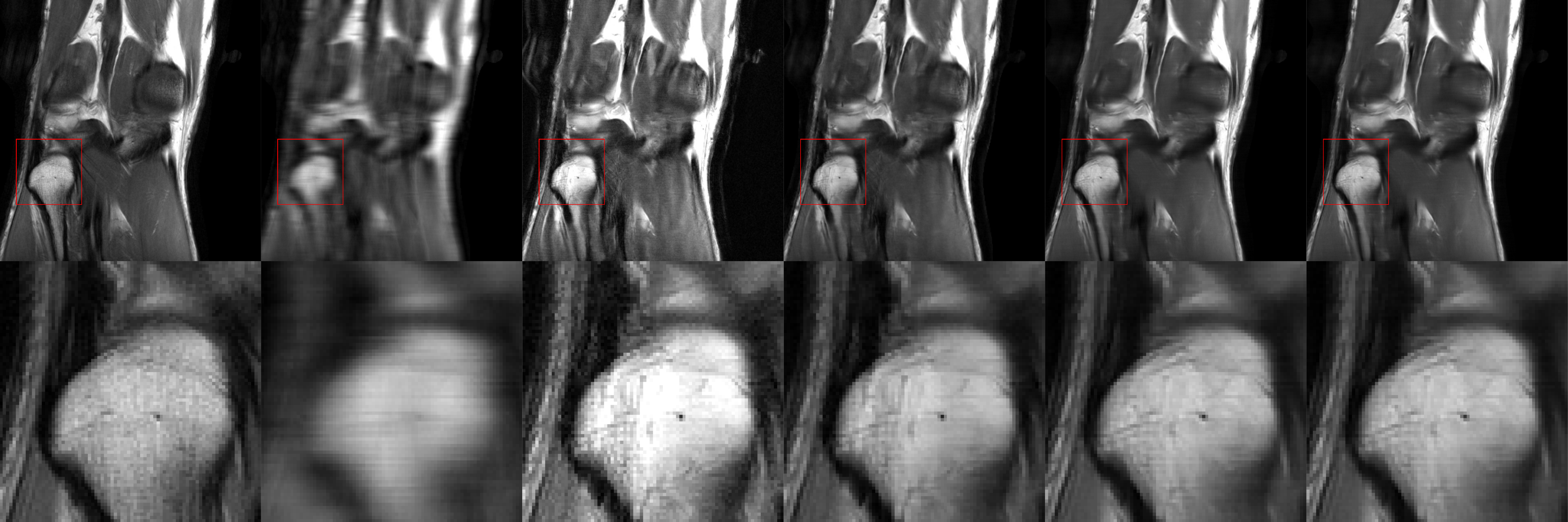}}
		\footnotesize (b)
	\end{minipage}
	\caption{(a) The reconstruction results on fastMRI dataset at four-fold acceleration. (b) The reconstruction results on fastMRI dataset at six-fold acceleration. The first rows of each subfigure are the reconstructed images and the second rows are zoomed details in the red squares. The corresponding images from left to right: the ground truth, zero-filled measurements, PGI, DGI, and the final output of $x_T^K$ and $x_z^K$ branch. \label{fig:6}}
\end{figure*}

\begin{table*}[!t]
	\caption{Quantitative results of the selected images at four-fold and six-fold acceleration rates.}\label{tab:4}
	\centering
	\resizebox{1.\columnwidth}{!}{
		\begin{tabular}{cccccc} 
			\toprule
			\multirow{2}{*}{Method} & \multirow{2}{*}{Image}  & \multicolumn{2}{c}{PSNR} & \multicolumn{2}{c}{SSIM} \\
			& & 4× & 6× & 4× & 6× \\
			\hline 
			\multirow{5}{*}{original, Coronal PD} & zero-filled  & 31.3089$\pm$3.3939 & 30.5176$\pm$3.3902 & 0.8778$\pm$0.0713 & 0.8626$\pm$0.0799 \\
			& $x_T$ & 31.3478$\pm$2.6593 & 29.7293$\pm$5.1889 & 0.4730$\pm$0.1556 & 0.6067$\pm$0.1215 \\
			& $x_T^0$ & 38.9772$\pm$3.6659 & 37.0103$\pm$4.1097 & 0.9508$\pm$0.0798 & 0.9370$\pm$0.0812\\
			& $x_T^{K}$ & 41.1457$\pm$3.6320 & 38.2893$\pm$3.9790 & 0.9630$\pm$0.0750 & 0.9450$\pm$0.0828\\
			& $x_z^{K}$ & \textbf{41.1981$\pm$3.3802} & \textbf{38.4491$\pm$3.2166} & \textbf{0.9634$\pm$0.0682} & \textbf{0.9452$\pm$0.0790}\\
			\hline
			\multirow{5}{*}{20\% sampling steps, Coronal PD} & zero-filled  & 31.3089$\pm$3.3939 & 30.5176$\pm$3.3902 & 0.8778$\pm$0.0713 & 0.8626$\pm$0.0799 \\
			& $x_T$ & 22.5488$\pm$7.6208 & 20.6821$\pm$7.8592 & 0.4664$\pm$0.2049 &  0.4601$\pm$0.2810 \\
			& $x_T^0$ & 37.8958$\pm$4.2813 & 36.6910$\pm$4.1992 & 0.9461$\pm$0.0809 & 0.9326$\pm$0.0837\\
			& $x_T^{K}$ & 40.9129$\pm$4.2478 & 38.2178$\pm$3.8602 & 0.9622$\pm$0.0817 & 0.9403$\pm$0.0826\\
			& $x_z^{K}$ & \textbf{40.9847$\pm$3.6107} & \textbf{38.2610$\pm$3.6829} & \textbf{0.9628$\pm$0.0770} & \textbf{0.9412$\pm$0.0798}\\
			\hline
			\multirow{5}{*}{original, FastMRI} & zero-filled & 33.2828$\pm$3.0594 & 32.5069$\pm$3.0657 & 0.8857$\pm$0.0709 & 0.8680$\pm$0.0791\\
			& $x_T$ & 26.5182$\pm$5.1008 & 23.9689$\pm$5.0221  & 0.4081$\pm$0.1377  & 0.4533$\pm$0.1307 \\
			& $x_T^0$ & 37.3427$\pm$3.5911 & 35.0240$\pm$3.3366 & 0.9236$\pm$0.0682 & 0.8933$\pm$0.0799\\
			& $x_T^{K}$ & 38.7848$\pm$3.4290 & 36.8528$\pm$3.0269 & 0.9338$\pm$0.0591 & 0.9092$\pm$0.0705\\
			& $x_z^{K}$ & \textbf{38.8241$\pm$3.3808} & \textbf{36.8730$\pm$3.0364} & \textbf{0.9344$\pm$0.0588} & \textbf{0.9094$\pm$0.0707}\\
			\bottomrule 
	\end{tabular}}
\end{table*}

\subsection{Evaluating the effectiveness of the proposed workflow}
To evaluate the effectiveness of the proposed workflow, we plotted the following images: the ground truth, zero-filled measurements, PGI ($x_T$), DGI ($x_T^0$), and the final output of $x_z^k$ and $x_T^k$ branches ($x_T^{K}$ and $x_z^{K}$) of an MR image from the test dataset.
The reconstructed images from the original sampling method at four-fold acceleration are shown in Fig.~\ref{fig:5} (a).
The first row shows the corresponding images, and the second row shows the magnified details. 
We can observe that the zero-filled image (second column) is blurred, and the PGI (third column) exhibits hallucination artifacts. 
However, the artifacts in the DGI (fourth column) were effectively eliminated by DMs from the zero-filled image and PGI.
Then, the GICs further enhanced the image quality of the DGIs, and $x_T^{K}$ (fifth column) and $x_z^{K}$ (sixth column) presented satisfactory results when compared with the ground truth images (first column), demonstrating the proposed coarse-to-fine workflow has powerful ability in recovering fine details.
Besides, we sampled PGIs from the pretrained score network, but using only 20\% of the original sampling steps.
The reconstructed images with 20\% sampling steps at four-fold acceleration are depicted in Fig.~\ref{fig:5} (b).
Adopting PGIs with worse quality, the proposed method still robustly achieve high-quality results, which are also consistent with the observations in Fig.~\ref{fig:5} (a).
Besides, we plotted reconstructed images from the FastMRI dataset at four-fold and six-fold in Fig.~\ref{fig:6} (a) and (b).
We observed unexpected artifacts on the PGIs, which lower the image quality.
However, through the denoising steps, we obseved that the artifacts have been greatly suppressed in DGIs, and the outputs of the unrolled network further eliminated the influence of noise and artifacts.

The quantitative results for the abovementioned methods are presented in Table \ref{tab:4}.
$x_z^{K}$, which is chosen as the final result, slightly outperforms $x_T^{K}$, and the other results obtained at four- or six-fold acceleration are consistent with the results observed from the images. 
Significantly reducing the sampling steps leads to further deterioration in the quality of SMLD results. However, the impact on the final reconstruction results is minimal, demonstrating the robustness of our proposed method to noise input, and the DMs are effective to extract valuable information from the noise input.
In summary, this experiment vividly illustrates how the coarse-to-fine workflow leverages SMLD results to extract crucial information, aiding in the reconstruction process and finally achieving high-quality results.

\subsection{Evaluating the network robustness with limited training slices}
\label{ltd}
\begin{figure*}[!t]	
	\begin{minipage}[b]{1\linewidth}
		\centering
		\centerline{\includegraphics[width=13.5 cm]{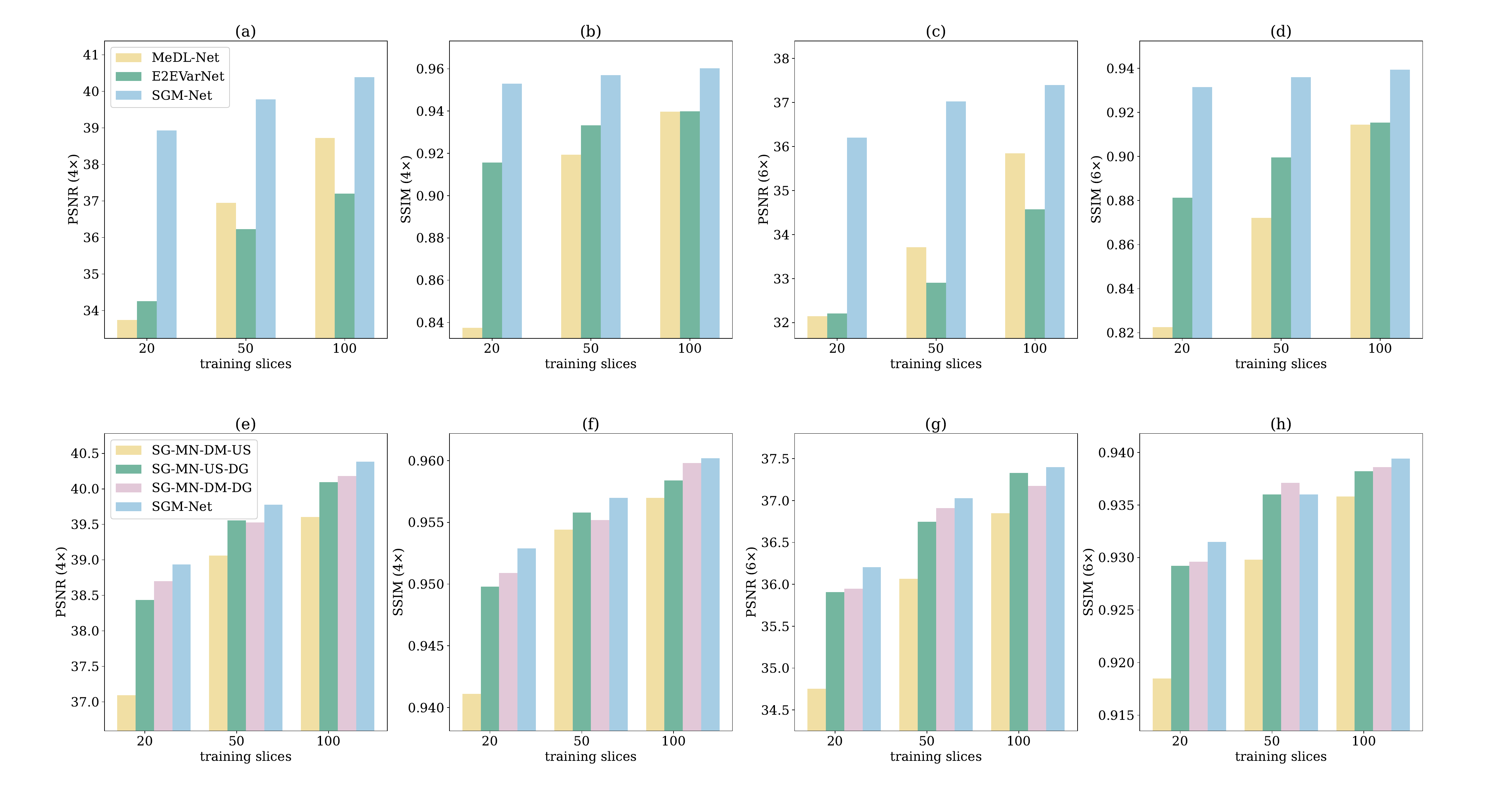}}
	\end{minipage}
	\caption{Four- and six-fold PSNR and SSIM results of different networks with limited training data.\label{fig:7}}
\end{figure*}

In this subsection we evaluated the network robustness with limited training data.
Here, we first compared with E2E-VarNet and MeDL-Net, which have shown convincing performances without utilizing guidance images.
We compared them with SGM-Net on the coronal PD sequence, and the PSNR and SSIM results at four-fold and six-fold are depicted in Fig~\ref{ltd} (a)-(d).
We trained the networks with 20, 50 and 100 slices, which are randomly selected, to evaluate their robustness.
We observed that the performances of E2E-VarNet and MeDL-Net deteriorated with decreasing training data.
Meanwhile, the performance of SGM-Net was more stable with greatly reduced training slices.

We also compared with three ablation networks (\textit{i.e.} SG-MN-DM-US, SG-MN-MN-DG and SG-MN-US-DG), and the results are depicted in Fig~\ref{ltd} (e)-(h).
Compared with E2E-VarNet and MeDL-Net, the ablation networks similarly showed better robustness, with a slower degradation in image quality under limited training data conditions, which also indicates the robustness of the proposed workflow.
Furthermore, the degradation of image quality become more pronounced with the decrease of training data, which demonstrated the necessity of the proposed modules.
To sum up, this experiment demonstrated that proposed guided reconstruction process can effectively compensate for information in the missing training data and achieve convincing performance robustly.

\section{Discussions}
To make the sampling process robust and achieve better performance, most of the cutting-edge methods proposed to optimize the training of the generative models or the sampling steps.
For example, the authors in \cite{quan2021homotopic} fed high-dimensional tensors to a noise conditional score network during training, and they designed homotopic iterations at the inference phase.
The authors in \cite{chung2022score} proposed a framework named score-POCS.
During inference, the reconstruction process iterates between SDE solvers and data consistency steps.
In \cite{ozturkler2023smrd}, the authors enhanced the reconstruction robustness of diffusion models with test-time hyperparameter tuning based on Stein's unbiased risk estimator.
Compared with them, the most significant distinction in our workflow is that we bypassed all the tedious hyperparameter tuning and network retraining steps and implemented a naive sampling. 
The reconstruction performances are then not robust especially with distribution-shift test datasets, and the results are observed to be severely corrupted.
We treat the SMLD results as a generalized prior, and we designed an unrolled network which can extract valuable information from the SMLD samples to guide reconstruction.
Experimental results have demonstrated that compared with the naive sampling results, only small amounts of data are required to train the unrolled network and the performances can be greatly improved.

Another bottleneck in diffusion models is that the sampling process is relatively slow.
To solve this problem, the authors in \cite{peng2022towards}, for example,  proposed DiffuseRecon, and the reverse process can be guided with observed k-space signal.
The authors designed a coarse-to-fine algorithm to save sampling time.
In \cite{hou2023fast}, the authors proposed deep ensemble denoisers and incorporated them with score-based generative models to reconstructed images.
Consequently, the necessary iterations are reduced.
In our workflow, we preserved the naive sampling process.
However, the unrolled network have good tolerance to artifact-affected inputs, hence the sampling steps can also be great reduced. 

In the proposed reconstruction workflow, the coarse SMLD samples have been demonstrated to be effective to guide the network training and improve the final results. 
Meanwhile, the proposed workflow is also compatible to optimized SMLD steps with higher quality references, which show great potential to be exploited in future works.
Besides, a limitation of the proposed method is that although the workflow exhibits considerable tolerance for the SMLD process, the unrolled network still requires some data for training to achieve superior performance. Training strategies such as self-supervising should be considered to further optimize the workflow in future works.

\section{Conclusions}
In this study, we proposed a novel MRI reconstruction workflow consisting of sampling, denoising and guided reconstruction steps.
Naive SMLD samples are denoised in DMs and the artifacts are effectively mitigated.
Then, the denoised samples are periodically updated and utilized to guide the network training process to recover fine details.
Experimental results demonstrated that the proposed method outperforms the workflows solely based on diffusion models or unrolled networks. 
Additionally, our method alleviates the substantial requirements for sampling steps and training images.
When compared with cutting-edge methods, our network consistently produces reconstructions of higher quality, showcasing its superiority in MRI reconstruction tasks.

\section{Acknowledgements}
This work was supported by the National Natural Science Foundation of China [Nos. 62331008,  62027827, 62221005 and 62276040], National Science Foundation of Chongqing [Nos. 2023NSCQ-LZX0045 and CSTB2022NSCQ-MSX0436], Chongqing University of Posts and Telecommunications ph.D. Innovative Talents Project [BYJS202211].

\bibliographystyle{elsarticle-num-names}  
\bibliography{Reference}
\end{document}